\documentclass[11pt,letterpaper]{article} 
\usepackage{cogsys}
\usepackage[T1]{fontenc}
\usepackage{times}
\usepackage{inconsolata}
\usepackage{graphicx}
\usepackage[font=small,labelfont=it,labelsep=period]{caption}
\usepackage{subcaption}
\usepackage{multirow}
\usepackage{color}
\usepackage{amsmath}
\usepackage{enumitem}
\usepackage{tabularx}
\usepackage{wrapfig}
\usepackage{enumitem}
\setitemize[0]{leftmargin=*,itemsep=-0.8pt}
\setenumerate[0]{itemsep=-1pt,leftmargin=*}
\usepackage{float}
\floatstyle{boxed}
\newfloat{myquote}{thp}{lop}
\floatname{myquote}{myquote}
\usepackage[backend=biber, style=apa, natbib]{biblatex}
\cogsysheading{3}{2014}{**--**}{9/2013}{5/2014}

\ShortHeadings{Indexical Model of Situated Comprehension}
              {S.\ Mohan, A.\ H.\ Mininger, and J.\ E.\ Laird}

\begin{document}

\title{Towards an Indexical Model of Situated Language Comprehension
 \\ for Cognitive Agents in Physical Worlds}

\author{Shiwali Mohan}{shiwali@umich.edu}

\author{Aaron H. Mininger}{mininger@umich.edu}

\author{John E. Laird}{laird@umich.edu}

\address{Division of Computer Science and Engineering, University of Michigan, Ann
  Arbor, MI 48109}
\vskip 0.2in

\begin{abstract}
We propose a computational model of situated language comprehension
based on the Indexical Hypothesis that generates meaning
representations by translating amodal linguistic symbols to modal
representations of beliefs, knowledge, and experience external to the
linguistic system. This Indexical Model incorporates multiple
information sources including perceptions, domain knowledge, and
short-term and long-term experiences during comprehension. We show
that exploiting diverse information sources can alleviate ambiguities
that arise from contextual use of underspecific referring expressions
and unexpressed argument alternations of verbs. The model is being
used to support linguistic interactions in Rosie, an agent implemented
in Soar that learns from instruction.

\end{abstract}

\section{Introduction}
\label{introduction}
As computational agents become pervasive in human society as
collaborators, the challenge of supporting flexible human-agent
interaction is becoming increasingly important. Natural language has
emerged as a strong contender for the modality of human-agent
interaction, as it is the primary means of communication in human
societies. Recent research in the design of interactive, intelligent
agents has shown that linguistic interaction is not only useful in
collaboration for human-agent tasks \citep{Fong2003,Kollar2010}, but it
also facilitates novel concept acquisition in interactive agents
\citep{Cantrell2011,Tellex2011,Mohan2012}. This has motivated research
on comprehensive models of natural language for collaborative task
execution and learning in human-agent teams.

\subsection{Situated Communication for Joint Activity}
In a joint activity, the speaker and the hearer pursue diverse
communicative goals in order to make progress on a task. Various types
of utterances are employed for expressing the communicative
goals. Imperative sentences such as \emph{Take out the trash} convey
that the speaker intends the hearer to complete a task. The joint
communicative goal is for the hearer to identify the intended task and
relevant objects and to correctly instantiate the task goals.  Other
utterances, such as assertions (\emph{Rice is in the pantry}), may be
used to establish shared beliefs about the state for joint
activity. Questions (\emph{Where is the milk?}) may be used to
supplement perceptual information by relying on the collaborative
partner's knowledge.

Communication between collaborators who are simultaneously embedded in
a shared task is \emph{situated}. The speaker's linguistic utterances
refer to objects, spatial configurations, and actions in the shared
environment.  To respond and react to utterances, the hearer must
associate the amodal linguistic symbols (\emph{words}) and
constructions (\emph{phrases}) with the modal representations of
perceptions, state, domain knowledge, goals, and policies that are
required for reasoning about and manipulating the environment.

Being situated provides a common ground of shared perceptions, goals,
and domain knowledge that can be exploited during linguistic
communication.  Information that is apparent from the current state of
the environment or that is a component of shared beliefs can be left
out of the linguistic utterance by the speaker. This results in more
efficient (fewer words) but ambiguous utterances. Humans frequently
use referring expressions such as \emph{it} or \emph{that cylinder}
that do not by themselves provide enough discriminative information
for unambiguous resolution. The speaker assumes that the hearer can
exploit extra-linguistic information, such as the context of the
ongoing discourse, for unambiguous comprehension.  Certain imperative
sentences such as \emph{take out the trash} incompletely specify the
action by omitting information such as the location where the
\emph{trash} should be moved. Such ambiguities
make situated comprehension a significant challenge for interactive
agents.

\subsection{Contribution and Claims}
In this paper, we study the utility of the Indexical Hypothesis
\citep{Glenberg1999} in developing comprehension models for
collaborative agents. These agents are embedded in physical tasks
that require the use of complex representations for probabilistic
perceptual processing, continuous spatial reasoning, and goal-oriented
task execution. To support situated communication, the comprehension
model must not only perform syntactical analyses, but also synthesize
meaning representations by associating linguistic information with
representations in other cognitive modules. \looseness=-1

The Indexical Hypothesis of language comprehension explains how
sentences become meaningful through grounding their interpretation in
situated action. The hypothesis asserts that comprehending a sentence
requires three processes: \emph{indexing} words and phrases to
referents that establishes the contents of the linguistic input,
\emph{deriving} affordances from these referents, and \emph{meshing}
these affordances under the guidance of physical constraints along
with the constraints provided by the syntax of the sentence. According
to the hypothesis, the linguistic information specifies the situation
by identifying which components (objects, relationships, etc.) are
relevant, and the semantic and experiential knowledge associated with
these components augments the linguistic input with details that are
required for reasoning and taking action. In this formulation of
language comprehension, linguistic symbols (words) and constructions
(grammatical units) are cues to the hearer to search their
perceptions, domain knowledge, and long and short-term experiences to
identify the referents intended by the speaker and to compose them.

Earlier work on the Indexical Hypothesis of language comprehension
identifies the processes that humans use for comprehension
\citep{Glenberg1999} and provides supporting data from human studies
\citep{Kaschak2000}. It does not describe the knowledge representations
and computational processes required for implementation of such models
on intelligent agents. The contribution of our work
is a computational model which we call the Indexical Model of
comprehension that precisely defines the representations and processes
described in the Indexical Hypothesis. We make two main claims:
\begin{enumerate}
\item The Indexical Model for comprehension can be used effectively by
  agents that act and learn in physical environments. We support this
  claim through demonstration: we describe an implementation of the
  model for Rosie (RObotic Soar Instructable Entity), an agent
  \citep{Mohan2012} that learns about various aspects of its
  environment through instruction.
\item The Indexical Model exploits diverse knowledge and experience of
  the domain to address ambiguity in semantic interpretation of the
  linguistic input. We evaluate this claim by demonstrating the
  utility of incorporating knowledge and experience in language
  comprehension on two ubiquitous problems - referring expression
  resolution and unexpressed argument alternation of verbs.
\end{enumerate}
\noindent
The rest of the paper is organized as follows. Section \ref{domain}
provides a description of our robotic domain and a brief overview of
Rosie. Section \ref{model-design} describes the indexical model. In
section \ref{analysis}, we explain how the model addresses
complexities that arise from ambiguity in its linguistic
input. Section \ref{related-work} discusses the related work on
designing comprehension models for agents. Section \ref{conclusion}
summarizes the paper and identifies our plans for future research.

\section {System Overview}
\label{domain}
Rosie is a instructable agent implemented in the Soar
cognitive architecture \citep{Laird2012}. It is embodied as a robotic
arm that can manipulate small foam blocks on a table top. The
workspace also contains four named locations: \emph{pantry},
\emph{garbage}, \emph{table}, and \emph{stove}. These locations
have associated simulated functionalities. For example, a \emph{stove}
can be turned on and off, and the \emph{pantry} can be opened and
closed. These functionalities change the state of the world. For
example, when the \emph{stove} is turned on, it changes the simulated
state of an object on it to \emph{cooked}. 

\subsection{Perception, Actuation, and Interaction}
Rosie senses the environment through a Kinect camera sensor.  The
perception system segments the scene into objects and extracts
features for three perceptual properties: color, shape, and
size. These properties along with the position and bounding volume of
the objects in the world are provided to Rosie, which uses them for
perceptual and spatial reasoning. For locations and objects, the
simulated state (such as \emph{open}, \emph{on}, \emph{cooked}) is
also included in its description.

To act in the world, Rosie sends discrete primitive commands to the
controller. These include object manipulation: \texttt{pointTo(obj)},
\texttt{pickUp(obj)}, and \texttt{putDown(x,y)}; and simulated
location operation: \texttt{open(loc)}, \texttt{close(loc)},
\texttt{turnOn(stove)}, and \texttt{turnOff(stove)}. The robot
controller converts these discrete commands to continuous closed loop
policies, which change the state of the environment. Human instructors
can interact with Rosie through a simple chat interface. Instructor's
utterances are preprocessed using the Link-Grammar parser
\citep{Sleator1991} to extract parts of speech and syntax. Rosie
responds using semantic structures that are translated to language
using templates. The instructor can point to an object by clicking on
it in the camera feed.

\subsection{Learning with Instruction}
Rosie begins with procedural knowledge for parsing language,
maintaining interactions, and learning from instruction. It also knows
how to perform primitive actions in the world. Through situated
interactive instruction, Rosie learns a novel word by acquiring a
corresponding concept and building an association between them. For
adjectives and nouns (such as \emph{red}, \emph{large}, or
\emph{cylinder}), the agent learns new classifications of perceptual
features (color, size, and shape) from interactive training. For
prepositions (such as \emph{right of}), the agent learns compositions
of primitive spatial predicates. For verbs (such as \emph{move}), the
agent learns novel task knowledge.

Learning in Rosie is interactive. Whenever it encounters a new word
that it cannot comprehend by associating it with known concepts, it
initiates interactions to learn the concept and the grounding of the
word. The interactions are situated in the environment. Through
instruction, the mentor provides specific examples of the concepts in
the environment. When the instruction is complete, Rosie can
comprehend the word and use the associated concept for classification,
spatial reasoning, and action. As the human-agent interaction is
linguistic, ambiguities may arise during instruction. A common form of
ambiguity arises due to the use of underspecific referring expressions
such as \emph{it} or \emph{that object}. Other ambiguities arise from
imprecise description of actions such as \emph{move the red cylinder
  to the table} that do not identify what relationship should be
established between the \emph{red cylinder} and the
\emph{table}. Rosie's comprehension model must alleviate such
ambiguities by incorporating information from its state perceptions,
domain knowledge, and experience.

\subsection{Memories and Their Contents}
We now give a brief description of the representations used in
Rosie. Detailed explanations can be found in our earlier work
\citep{Mohan2012}. The agent's beliefs about the current state are held
in its working memory. These beliefs are \emph{object-oriented} and
are derived from its perceptions of the world, from its experiential
knowledge of the world encoded in its long-term memory, and from its
interactions with the human collaborator.

Rosie's visual knowledge is encoded in its long-term perceptual
memory. The memory accumulates training examples that are used to
classify objects in terms of visual attributes: color, size, and
shape. Each visual attribute has a k-nearest neighbor (kNN) classifier
associated with it and each class within the kNN is referred to using
a perceptual symbol. For example, the domain of the \emph{color}
attribute may contain perceptual symbols {\texttt{C22}, \texttt{C53},
  \texttt{C49}}, each of which correspond to colors known to
Rosie. All perceptible objects are represented in working memory along
with the known value assignments to their visual attributes.

Rosie's spatial knowledge is distributed between its semantic memory
and its spatial-visual system. It learns and represents spatial
prepositions such as \emph{on} and \emph{near} as compositions of
known primitive spatial literals in the spatial-visual system that
encode alignment along axes and distance between objects. It generates
symbolic representations of spatial relationships between perceptible
objects using this knowledge. This representation is useful for
reasoning about existing spatial relationships on the workspace (such
as \emph{the red cylinder is on the stove}) and executing actions that
establish specific spatial relationships between arguments (such as
\emph{put the red cylinder on the stove}).

Rosie can learn goal-oriented tasks, such as \emph{cook a steak}, that
require it to achieve a composition of spatial and state predicates by
executing a policy defined hierarchically over primitive actions.  Its
task knowledge is distributed across its semantic and procedural
memories. While Rosie's semantic memory stores a task-concept network
that includes the goal definition of the task and constraints on how
the goal should be instantiated, its procedural memory encodes the
task's availability conditions, policy, and termination conditions
represented as rules and implemented through operator proposal,
selection, and application.

\section{The Indexical Model of Comprehension}
\label{model-design}
Consider the imperative sentence \emph{move the large red cylinder to
  right of the blue triangle}.  We assume that the speaker makes this
utterance because she intends for the hearer to execute the requested
action. The process of indexical comprehension involves identifying
the referents of the linguistic input and composing them to generate
an action instantiation that is grounded in the modal symbols that
support reasoning about and manipulation of the environment. Following
the Indexical Hypothesis, comprehension is carried out in three stages
described below. Note that in the remainder of this paper, the
impelmentation of the Indexical Model is referred to as \emph{the
  model} and the complete agent that includes the perception,
actuation, and learning components along with the Indexical Model for
comprehension is referred to as Rosie.

\subsection{Indexing}
\label{indexing}
\begin{figure}[t]
\vskip 0.05in
\begin{center}
\includegraphics[width=0.95\textwidth]{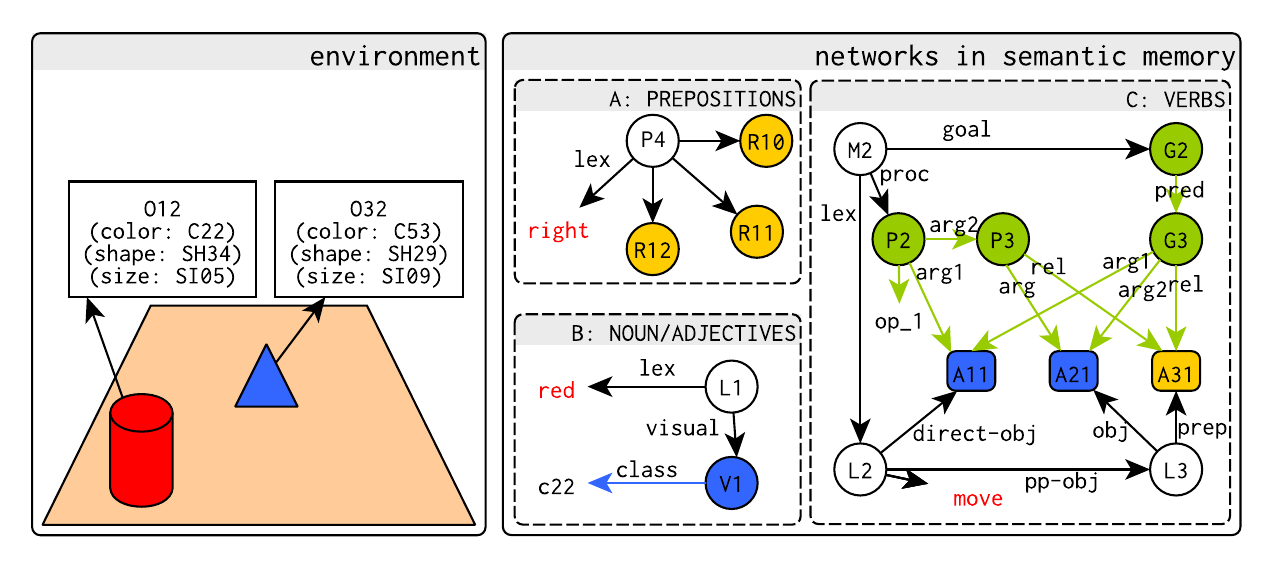}
\caption{Environment state and the knowledge encoded in Rosie's
  semantic memory. The white nodes (\texttt{P4, L1, M2, L2, L3})
  represent indexical maps between amodal linguistic symbols (in red:
  \texttt{right, red, move}) and modal domain knowledge. Yellow nodes
  (\texttt{R10, R11, R12, A31}) represent spatial symbols and slots
  (round rectangles: \texttt{A31}), blue nodes (\texttt{V1, A11, A21})
  represent visual symbol and slots, and green nodes (\texttt{P2, P3,
    G2, G3}) represent procedural symbols.}
\label{fig:indexical-maps}
\end{center}
\vskip -0.2in
\end{figure}

After preliminary lexical processing, it is established that the
linguistic input contains two referring expressions (REs: \emph{the red
  cylinder} and \emph{the blue triangle}), a spatial preposition
(\emph{to the right of}), and a verb (\emph{move}). The goal of the
\emph{indexing} step is to identify the referents for these linguistic
units. The model uses a simple referential grammar: nouns and
adjectives refer to visual properties; referring expressions refer to
objects; prepositions refer to spatial relationships; and verbs refer
to tasks. Figure \ref{fig:indexical-maps} shows the objects
(\texttt{O12}, \texttt{O32}) and semantic networks \texttt{A},
\texttt{B}, and \texttt{C} that form the referent set of REs,
prepositions, nouns/adjectives, and verbs. We introduce the term
\emph{indexical maps} for structures in semantic memory that encode
how linguistic symbols (nouns/adjectives, spatial prepositions, and
action verbs) are associated with perceptual symbols, spatial
compositions, and action-concept networks. Now we describe how
indexical maps are used during comprehension. In the following
text, $R^{super}_{sub}$ denotes the referent set. The
superscript $super$ denotes the contents of the set ($o$ for objects,
$s$ for spatial relations, and $a$ for actions/tasks) and the
subscript $sub$ denotes the words used to generate the set. 
 
To index REs (\emph{the red cylinder}), the model must first index the
descriptive words (\emph{red} and \emph{cylinder}). For each of these
words, the model queries the semantic memory for a node that was
previously learned to be associated with the lexical string. For the
string \emph{red}, the memory returns node \texttt{L1} (refer to
Figure \ref{fig:indexical-maps}). Node \texttt{L1} \emph{maps} the
lexical string \emph{red} to the corresponding perceptual symbol
\texttt{C22} which is a class in the color classifier. Once the model
has retrieved perceptual symbols for all words, it searches working
memory for objects that have the required perceptual symbols. These
objects are assumed to be the intended referents of the RE. In cases
where the RE is underspecific (e.g., \emph{this block}), there may be
multiple objects that match, resulting in ambiguity. The model can use
other kinds of information to resolve such ambiguities as we describe
in Section \ref{rresolution}. For the sake of simplicity, in this
example we assume that only one object (\texttt{O12}) matches the
cue. This object is included in the referent set ($R^o_{red, cylinder}=
\texttt{O12}$) for the RE \emph{the red cylinder}. Similarly,
$R^o_{blue, triangle} = \{ \texttt{O32}\}$ for the RE \emph{the blue
  triangle}. If these sets are empty, it indicates that Rosie lacks
knowledge to generate groundings of the RE, in which case it prompts
the instructor for training examples.

Prepositions are indexed in a similar fashion. For a preposition
string (\emph{right}), the model queries the semantic memory for an
indexical node that had previously been learned and
is associated with it. On the retrieval of the
requested node (\emph{P4}), the model creates the referent set
$R^s_{right} = \{\texttt{P4}\}$. If the set is empty, Rosie asks the
instructor to provide an example of \emph{right-of} in the
environment.

To index the verb \emph{move}, the model queries the semantic memory
for a node that is connected to the string \emph{move}. The memory
returns the node \texttt{L2}. Then the model retrieves the mapping
node \texttt{M2} that associates the verb to domain knowledge of the
task - the goal definition \texttt{G2} and procedural operator node
\texttt{P2}. The referent set for the verb consists of the
task-concept network, $R^a_{move} = \{\texttt{(M2,P2,G2)}\}$.

\subsection{Instantiating Domain Knowledge}
Once the referents have been identified, the next step is to retrieve
the domain knowledge associated with them and instantiate it under the
syntactical constraints of the sentence. The model begins by
retrieving the previously learned syntactical nodes associated with
the verb \emph{move}. The sentence \emph{move the large red cylinder
  to the right of the blue triangle} has a direct object (RE,
\emph{the large red cylinder}) and a prepositional object (RE,
\emph{the blue triangle}) connected to the verb through the
preposition \emph{right}. Following this syntactical structure, the
model retrieves the direct object (\texttt{direct-obj} in the figure)
node \texttt{A11} and the prepositional phrase object
(\texttt{pp-object} in the figure) node \texttt{L3} which is further
expanded to retrieve nodes \texttt{A21} and \texttt{A31}. The slot
nodes \texttt{A11} and \texttt{A21} can receive sets of objects in the
environment. \texttt{A11} is filled by $R^o_{red, cylinder}$ as
\emph{the red cylinder} is the direct object of the verb \emph{put}
and \texttt{A21} is filled by $R^o_{blue, triangle}$, as \emph{the
  blue cylinder} is the RE in the prepositional phrase of the
verb. \texttt{A31} is a spatial slot that is filled by $R^s_{right}$,
the referent set for \emph{right}.

Next, the model expands the domain knowledge nodes \texttt{P2} and
\texttt{G2}. The subgraph (\texttt{P2, P3, A11, A21, A31}) constrains
how the policy operator \texttt{op\_1} is instantiated. The subgraph
(\texttt{G2, G3, A11, A21, A31}) constrains the instantiation of goal
of the task. The values of the slot nodes (\texttt{A11}, \texttt{A21},
\texttt{A31}) determine the contents of the goal and the policy
operator \texttt{op\_1}. Instantiation of domain knowledge results in
the interpretation set $I_s$, which contains elements that encode:
\texttt{execute} policy \texttt{op\_1} defined over objects drawn from
sets $R^o_{red, cylinder}$, $R^s_{right}$, and $R^o_{blue, triangle}$
until the corresponding goal is achieved. For the imperative
\emph{move the large red cylinder to the right of the blue triangle}
in the scene in Figure 1, the policy \texttt{op\_1} will be
instantiated over arguments \texttt{O12}, \texttt{O32}, and
\texttt{P3} with the goal \texttt{P3(O12,O32)}. The policy
\texttt{op\_1} selects \texttt{pick-up(O12)} if the arm is empty and
\texttt{put-down(O12,P3,O32)} if the arm is holding \texttt{O12}.

In \citet{Glenberg1999} formulation of the
Indexical Hypothesis, this step was described as \emph{deriving the
  affordances}. However, the term \emph{instantiating domain
  knowledge} better describes our formulation of the process.

\subsection{Meshing}
The interpretation set $I_s$ is the set of different groundings of the
imperative sentence, which can have several elements arising from
underconstraining cues in the linguistic input. However, only a subset
of these groundings can be executed in the environment given physical
constraints and spatial relationship between objects. For example, the
\emph{open} action can only be executed for \emph{stove} and
\emph{pantry}. When \emph{open} is used with an underconstraining RE
such as \emph{it}, there will multiple interpretations, but only two
of those interpretations can be executed in the environment.

Suppose $A$ is a set of tasks that can to be executed in the current
state based on their availability conditions. The intersection set
$I_s \cap A$ is the set of tasks that the instructor intends Rosie to
execute. If this set contains a single element, that task operator is
selected and executed. If this set contains multiple elements, further
interaction or internal reasoning is necessary for resolution. The
cardinality of the referent sets ($R$) is used to determine the source
of the ambiguity.  Rosie asks questions to gather information that
will reduce the cardinality of the ambiguous set. If the $I_s \cap A =
\phi$, Rosie does not have enough knowledge to generate the correct
groundings for the required task. This is an opportunity to learn the
task, so Rosie begins a learning interaction by prompting the human
collaborator to present an example execution.

\section{Dealing with Complexities}
\label{analysis}
Various issues can arise while attempting to generate an unambiguous and
complete interpretation of an utterance. Ambiguities occur when the
linguistic cues underspecify their referents, resulting in multiple
elements in their referent sets and, consequently, multiple
interpretations. In this section, we describe how the Indexical Model
deals with ambiguities that arise from the contextual use of referring
expressions and with situations in which the information required to
instantiate a policy is not completely specified in the linguistic
input.

\subsection{Reference Resolution}
\label{rresolution}
Humans use a variety of surface forms to refer to the same entity. A
few, such as definite noun phrases (\emph{the large red cylinder on
  the table}), may uniquely identify the intended referent from the
current shared perceptions. However, a majority of referring
expressions (REs) encountered in conversations, such as noun phrases
with indefinite determiners (\emph{a cylinder}), demonstrative/diectic
pronouns (\emph{this}, \emph{that}), and personal pronouns (\emph{it}),
are ambiguous.

For the generation and comprehension of REs, the communicative goal is the
identification of the intended object by the hearer. The form of REs
and other linguistic (word order) and phonetic (intonation) aspects
are influenced by the cooperative speaker's assumptions about the
relative salience of referents to the hearer. An object might be more
salient than others because it is useful in performing a task, it is
being pointed at, it changes appearance, or it is unexpected. The
ongoing discourse can also make objects more salient. Speakers make
assumptions about which objects are more salient to the hearers and
use these assumptions to choose an appropriate RE. More salient
objects can be referred to by less informative REs, as the hearer can
exploit saliency for disambiguation. \looseness=-1

\citet{Gundel1993} express the notion of the current and historical
salience of an object to the hearer as its \emph{cognitive
  status}. They propose a Givenness Hierarchy (GH) that relates the
cognitive status of objects with different RE surface forms. The GH
identifies six cognitive statuses, only four of which are relevant to
this paper: \emph{in-focus} (personal pronouns) $>$ \emph{activated}
(demonstrative pronouns, demonstrative noun phrases) $>$
\emph{uniquely-identifiable} (definite noun phrase) $>$
\emph{type-identifiable} (indefinite noun phrase). Each status in the
GH is the necessary and sufficient condition for use of the
corresponding RE and entails all the lower statuses. The choice of a
RE form by the speaker is indicative of what cognitive status is
useful for resolution. Given the cognitive status of an object and the
hearer's knowledge about the environment, the information in the RE
uniquely identifies the intended referent.

\subsubsection{Non-linguistic Contexts}
\label{context}
\citet{Knoeferle2006} identify two dimensions of the interaction
between the linguistic and situated context: \emph{informational} and
\emph{temporal}. The first dimension refers to the rapid integration
of diverse information for various cognitive modules including
perceptual processes and domain knowledge. The second dimension refers
to the temporal coordination between attentional processes and
utterance comprehension. While REs such as noun phrases (lower in the
GH) exploit the informational dimension of language-context
interaction, ambiguous REs such as pronouns (higher in the GH) exploit
the temporal dimension. To process the complete range of RE forms in
the GH, the Indexical Model exploits both the informational (described
earlier) and the temporal dimensions (described below).

\noindent
\textbf{\emph{Interaction}}: When conversation participants
communicate, they focus their attention on only a small portion of
what each of them perceives, knows, and believes. Some entities
(objects, relationships, actions) are central to information transfer
at a certain point in dialog and, hence, are more salient than
others. This is exploited by both the speaker and the hearer. It lets
the speaker refer to focused entities with minimal information and
lets the hearer heuristically constrain the set of possible
referents, reducing cognitive load on both.

Rosie has a model of instructional interaction \citep{Mohan2012} based
on the computational theory of task-oriented discourse by
\citet{Grosz1986}. The interaction model organizes the discourse
structure according to the goals of the task. The current state of the
human-agent interaction is represented by three
elements. \emph{Events} cause change, either in the environment
(\texttt{actions} such as \texttt{pick-up(O32)}), the dialog
(\texttt{utterances} such as \emph{Where is the red cylinder?}), or
Rosie's knowledge (\texttt{learning events} such as explanation-based
learning). A \emph{segment} is a contiguous sequence of events that
serves a specific \emph{purpose} and organizes the dialog into
purpose-oriented, hierarchical blocks in accordance with Rosie's
goals. The \emph{purpose} of a segment is determined based on
pre-encoded heuristics about instructional interactions. Finally, the
focus of the interaction is captured in a \emph{stack} of active
segments. When a new segment is created, it is pushed onto the
stack. The top segment of the stack influences the agent's processing
by suggesting a purpose that Rosie should act to achieve. When the
purpose of the top segment is achieved, it is popped from the stack.

The stack maintains a set of all referents (objects, spatial
predicates, actions) that are related to the events in its active
segments. The set of objects ($O^{stack}$) is most pertinent to this
paper because it identifies all the objects that have been referred to in
the current discourse, making them more salient than other perceivable
objects.

\noindent
\textbf{\emph{Attention}}: Object referents that have been brought to
attention, either through linguistic or extra-linguistic means, but
are not in the focus of the ongoing communication are usually referred
to by demonstrative pronouns or demonstrative noun phrases
(\emph{this}, \emph{that cylinder}) \citep{Gundel1993}. The
extra-linguistic means may include pointing by the speaker or
unexpected stimulus such as a loud noise. To resolve such REs, Rosie
must maintain the history of references to objects in its perceptions.

Rosie uses the architectural recency-based activation in Soar's
semantic memory as a form of attention. The recency-based activation
biases retrieval from semantic memory towards the more recently
accessed elements. An object is \emph{accessed} only if it was pointed
at or was used in an action or learning. Anytime an object is
accessed, Rosie stores its representation in the semantic memory which
boosts its activation in accordance with recency computation. A
completely ordered subset $O^{active}$ of the highest activated $n$
objects is retrieved from Rosie's semantic memory to its working
memory. These are combined with objects in focus to give a set of
objects to which Rosie is \emph{attending} ($O^{attend} = O^{stack}
\cup O^{active}$). This formulation of attention combines linguistic
and extra-linguistic notions of salience.

\subsubsection{Resolving References in the Indexical Model}
In Section \ref{indexing}, we described the indexing of referring
expressions in simple cases where the words in the RE and their
corresponding perceptual symbols by themselves uniquely identified the
referent object. Here, we give details about how an ambiguous RE is
indexed by incrementally adding diverse types of information in the
Indexical Model.
\vskip -1in
\begin{enumerate}[start=0]
\item \emph{Maintain cognitive status}. Following the Givenness
  Hierarchy, the model maintains different cognitive statuses for
  objects:
  \begin{itemize} [noitemsep,nolistsep]
    \item Objects in the interaction stack ($O^{stack}$) have the
      \emph{in-focus} status;
    \item Objects that are being attended to ($O^{active}$) have the
      \emph{activated} status;
    \item Objects in perceptions ($O^{percept}$) have the
      \emph{identifiable} status.
  \end{itemize}

\item \emph{Assign resolution type}. For any RE $r$, the model
  determines its resolution type based on its surface form. If the RE
  is:
  \begin{itemize} [noitemsep,nolistsep]
    \item a definite noun phrase (\emph{the red cylinder}),
      demonstrative pronoun (\emph{this}), or personal pronouns
      (\emph{it}), the speaker has a specific intended referent and
      comprehension should unambiguously determine it (\emph{unique}
      resolution);
    \item an indefinite noun phrase (\emph{a red cylinder}), this
      indicates that there is no specific intended referent and any
      object that fits the noun phrase can be used for resolution
      (\emph{any} resolution).
  \end{itemize}

\item \emph{Determine the candidate referent set}. The model exploits
  the heuristic that surface forms of REs indicate which set contains
  the intended referent. The candidate referent set is:
  \begin{itemize} [noitemsep,nolistsep]
    \item $R_r^o=O^{stack}$ for personal pronouns (\emph{it});
    \item $R_r^o=O^{attend}$ for demonstrative pronouns (\emph{this},
      \emph{that}) and noun phrases (\emph{this cylinder});
    \item $R_r^o=O^{percept}$ for definite (\emph{the cylinder}) and
      indefinite (\emph{a cylinder}) noun phrases.
  \end{itemize}

\item \emph{Apply the visual filter}. Rosie's knowledge of the
  perceptual symbols and how they relate to words is useful in
  identifying the referents of descriptive REs (\emph{the red
    cylinder}). The model indexes each descriptive word (\emph{red},
  \emph{cylinder}) in a noun phrase, and then the model looks up its
  corresponding perceptual symbols, which are collected into a set as a
  cue. All the objects in the candidate set ($R_r^o$) whose working
  memory representations do not contain this cue are deleted from this
  set.
  
\item \emph{Apply the spatial filter}. If the RE uses spatial
  reference (\emph{the cylinder on the right of the pantry}), referent
  sets for both noun phrases ($R^o_{cylinder}$, $R^o_{pantry}$) are
  obtained. The model indexes the preposition \emph{right} to
  retrieve the corresponding spatial relationship predicate
  \texttt{P4}. Items in $R^o_{cylinder}$ that do not satisfy the
  relationship \texttt{P3} with any item in $R^o_{pantry}$ are
  deleted. This is a meshing step that combines linguistic information
  with the domain knowledge and the perceptual state. \looseness=-1

\item \emph{Apply the task filter}. If the REs are used with verbs, as
  in an action command (\emph{put the cylinder in the pantry}), the
  model uses the knowledge of task restrictions to constrain their
  interpretation. To access this knowledge, the model indexes the verb
  to retrieve a task-operator and its corresponding goal. During
  meshing, it looks at all task-operator instantiations that are
  applicable in the current environmental state under the physical
  constraints and the knowledge of object affordances. Any object that
  does not occur in the arguments of currently applicable task
  instantiations is removed from $R_r^o$ of the RE.

\item \emph{Obtain partial ordering}. The elements of the referent
  set ($r \in R_r^o$) are partially ordered based on their cognitive
  status and resolution type. If resolution is \emph{unique}
  (from step 1), then $r_i \in O^{stack} > r_j \in O^{active} >
  r_k \in O^{percept}$. If resolution is \emph{any}, then all objects
  have equal preference. \looseness=-1

\item \emph{Resolve}. After applying all available filters, if $R_r^o$
  contains only a single object, that object is selected as the
  intended referent. If it contains multiple objects, the model uses
  the partial ordering obtained earlier to select the object highest
  in the order as the intended referent. If the partial ordering is
  not informative enough for resolution, the model initiates a
  subdialog to obtain more information from the instructor. If the
  resolution is \emph{any}, all objects have equal preference and one
  is chosen at random.
\end{enumerate}

The resolution process described here integrates seamlessly with the
incremental learning modules and the interaction model for
mixed-initiative conversations. 

\subsubsection{Evaluation and Analysis}
\begin{figure}[t!]
     \begin{subtable}[b]{0.55\textwidth}
       \vskip -0.15in
           \begin{footnotesize}
             \begin{subtable}[b]{1\textwidth}
               \begin{tabular}[b]{|p{3.05cm}|}
                 \hline
                 Sample Dialog \\
                 \hline
                 I: \emph{Pick it up.} \\
                 A: \emph{Which object?} \\
                 I: \emph{the blue one.} \\
                 A: \emph{Which blue object?} \\
                 I: \emph{the cylinder.} \\
                 A: \emph{Which blue cylinder?} \\
                 I: \emph{the one in the pantry.} \vspace{0.85em}
                 \\ \hline
               \end{tabular}
               \qquad \hspace{-0.75cm}
               \begin{tabular}[b]{|p{3.7cm}|}
                 \hline
                 Models \& Dimensions \\
                 \hline
                 \emph{p}: perceptual semantics\\
                 \emph{p+t}: perceptual semantics, task knowledge\\
                 \emph{p+t+a}: perceptual semantics, task knowledge, attention \\
                 \emph{p+t+a+d}: perceptual semantics, task knowledge, attention, dialog context\\ \hline
                 
               \end{tabular}
             \end{subtable}
          \begin{tabular}[b]{|cp{5.2cm}|}
            \hline
            \multicolumn{2}{| c |}{Scenario Ambiguity}\\ \hline
            Ambiguity 1 & only intended referents \\
            Ambiguity 2 & perceptually distinct distractors \\
            Ambiguity 3 & distractors: different shapes, same colors \\
            Ambiguity 4 & distractors: same colors and shapes \\ \hline
         \end{tabular}
       \end{footnotesize}
    \end{subtable}
    \qquad \hspace{-1.5cm}
    \begin{subfigure}[b]{0.5\textwidth}
    \includegraphics[width=\textwidth]{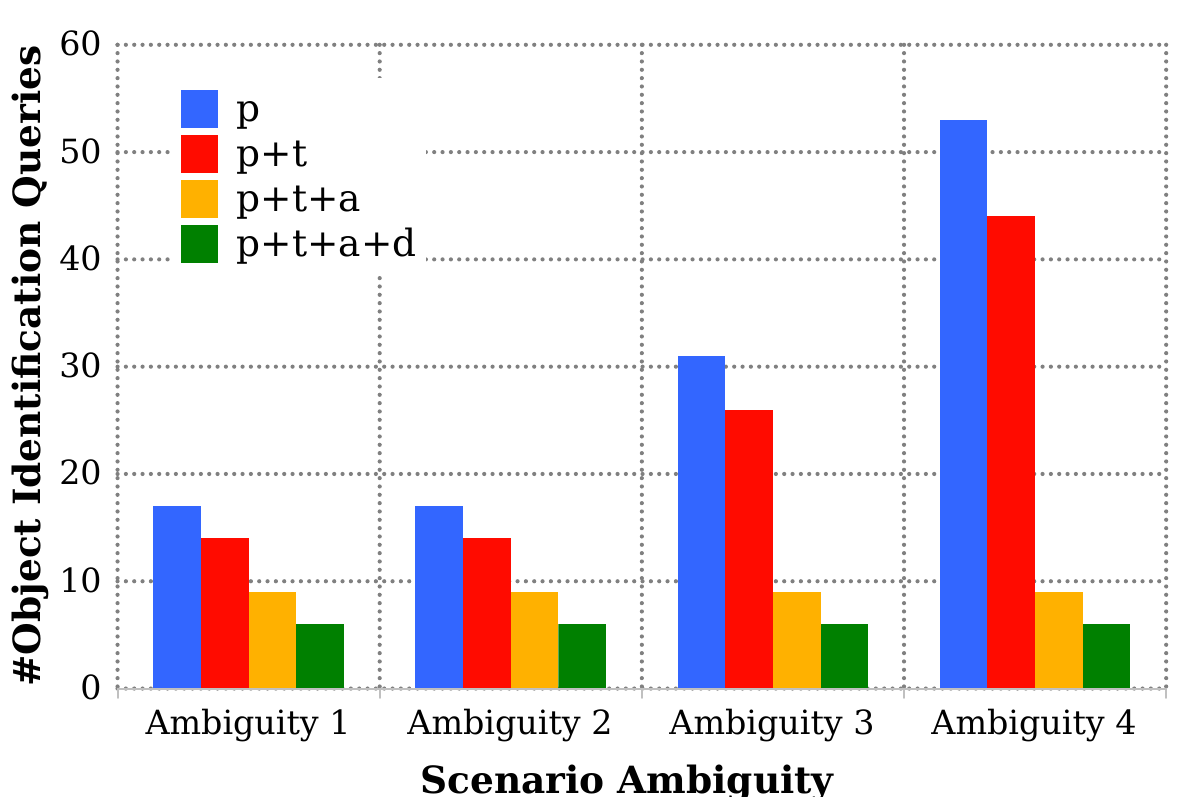}
    \end{subfigure}
    \caption{(left) A sample dialog and various models and
      scenarios used for evaluation of RE resolution. (right)
      Number of object queries asked by Rosie for RE resolution.}
    \label{fig:rresolution}
\vskip -0.2in
\end{figure}

\emph{\textbf{Experiments}}. We generated a corpus of 25 instructor
utterances that address different capabilities of Rosie. This corpus
contains instruction sequences that teach and query Rosie about
objects and their attributes, present and verify grounded examples of
spatial prepositions, and teach verbs. This corpus also contains
references to three objects in the scene that use varying forms of
referring expressions, including 12 instances of personal pronouns
(such as \emph{it}), four instances of demonstrative pronouns (such as
\emph{this}), three instances of demonstrative phrases (such as
\emph{that cylinder}), and 14 varying length noun phrases with
different descriptive words (such as \emph{the red cylinder}).  We
evaluated alternative models of comprehension that exploit the
informational and temporal dimensions of non-linguistic context. The
baseline model \emph{p} uses the context derived from perceptual
semantics only. Model \emph{p+t} exploits the restrictions derived
from task knowledge along with perceptual semantics while model
\emph{p+t+a} exploits the temporal dimension by encoding the
attentional state. Model \emph{p+t+a+d} encodes both the attentional
and dialog states. Each comprehension model was evaluated using the
instruction corpus on different scenarios of increasing perceptual
ambiguity in the environment obtained by adding distractor objects as
shown in Figure \ref{fig:rresolution}.

\noindent
\emph{\textbf{Evaluation metric}}. Rosie is an interactive agent that engages
the human instructor in a subdialog if it fails at any stage in its
processing. On failing to resolve ambiguous referring expressions in
sentences, Rosie asks questions to obtain more information that will
constrain its resolution. The instructor can, then, incrementally
provide more identifying information. An example dialog is shown in
Figure \ref{fig:rresolution}. The question-answer pairs (\emph{object
  identification} queries) are informative of how ambiguous an RE is
to the model given the ambiguity in the current scenario and the
contexts. Note that the instructor could have provided all the
identifying information in a single response (\emph{Which object?},
\emph{the blue cylinder in the pantry}). However, letting Rosie take
the initiative in resolution ensures that it accumulates the minimum
information required for unique identification in the current
situation. The number of \emph{object identification} queries in this
set up is positively correlated with the number of objects in $R^o_r$
after all filters have been applied.

\noindent
\emph{\textbf{Results}}. The graph in Figure \ref{fig:rresolution} shows the
number of \emph{object identification} queries asked by Rosie while
using different comprehension models in scenarios with varying
perceptual ambiguity. The models reliably integrate information
provided incrementally over several interactions for resolution.
Consequently, all REs were eventually correctly resolved in all models
in all scenarios.  The model \emph{p+t+a+d} can exploit the
informational and temporal dimensions effectively for resolution. To
establish that the non-linguistic context contributes information
above and beyond what is encoded in the linguistic features, we ran
Stanford CoreNLP \citep{Lee2012} on our corpus. Co-reference resolution
in CoreNLP failed to correctly resolve ten (28.6\%) references. These
results suggest that the grounded contexts are essential for robust
comprehension in an embodied agent.

The baseline model \emph{p}, which only exploits the contexts derived
from perceptual semantics, generates the most queries for all levels
of ambiguity. Model \emph{p+t} is able to use its knowledge about the
task to constrain resolution and, therefore, requires fewer queries
for achieving the same resolution results. The models that exploit
both the temporal and informational dimensions require even fewer
queries to achieve similar performance across all
scenarios. Conversing with agents that only encode the informational
dimension of non-linguistic context usually requires wordy REs, such as
\emph{the red cylinder in the pantry}, that must be repeated in all
interactions related to that object. The use of the temporal dimension
for comprehension allows the use of shorter referring expressions
(\emph{it}, \emph{this cylinder}), resulting in efficient
communication.

As perceptual ambiguity in the environment increases, models that
exploit only the informational dimension (\emph{p}, \emph{p+t})
require more perceptual information for resolving REs. Models that
exploit the temporal dimension (\emph{p+t+a}, \emph{p+t+a+d}) ask the
same number of queries across all scenarios, demonstrating that the
use of co-reference is an efficient way to communicate about objects
in human-agent dialogs. It lets the instructor communicate the
intended referent without incorporating large amounts of information
in utterances in perceptually ambiguous scenarios.

\subsection{Unexpressed Argument Alternations of Verbs}
\label{missing-object}
In Rosie, the goal of comprehension of an imperative sentence is to
correctly instantiate a task that can be executed in the
environment. The verb of the sentence identifies the task and the
verb's objects indentify the arguments of the task. The syntax is
useful in instantiating the task goals and a policy that can be
executed in the environment to achieve them. The syntax of English
verbs is flexible and often omits objects. \citet{Levin1993}
characterizes the variations in verb usage as
\emph{alternations}. Consider, for an example, an imperative
\emph{take the trash out to the curb} that informs the hearer that the
direct object \emph{trash} has to be placed on the location
\emph{curb}. An alternative imperative sentence that conveys the same
meaning is \emph{take the trash out}. The target location of the
\emph{trash} is left unexpressed. This alternation is termed
\emph{unexpressed object alternation} by \citet{Levin1993}. This verb
alternation pose a significant challenge to an agent that seeks a
precise action interpretation that can be executed in the environment.

Humans generate and comprehend such sentences by relying on the shared
knowledge about the domain. In the example, both the speaker and the
hearer know that the \emph{trash} is usually put on the
\emph{curb}. This lets the speaker omit the location in the sentence
\emph{take the trash out} for the sake of communicative
efficiency. The choice of this syntax by the speaker indicates that
they assume that the hearer can fill the missing location from their
knowledge of the domain. Upon hearing the utterance, the hearer must
exploit this knowledge and generate an appropriate, complete
representation of the action.

\subsubsection{Exploiting the Hearer's Instructional Experience}
To deal with imperative sentences with unexpressed information about
the action, the model relies on Rosie's prior experiences of
interacting with the instructor and acting in the domain. Consider the
verb \emph{move} and the variations of imperatives that can be
constructed from it:
\vspace{-0.2cm}
\begin{itemize}[leftmargin=0.8cm]
\item[(a)] \emph{Move the green object to the right of the table}.
\item[(b)] \emph{Move the green object to the table}. 
\end{itemize}
\vspace{-0.2cm} In (a), the direct-object \emph{the green object}, the
location \emph{the table}, and their spatial relationship (\emph{right
  of}) are completely specified. In (b), the spatial relationship is
omitted with an understanding that there is a default configuration
(\emph{on}) between the object and the location. We extend
\citet{Levin1993} characterization of
unexpressed object alternations to include other types of verb
arguments (spatial relationship in this case) and term it
\emph{unexpressed argument alternation}.

\begin{figure}[t!]
    \begin{subfigure}[b]{0.48\textwidth}
    \includegraphics[width=\textwidth]{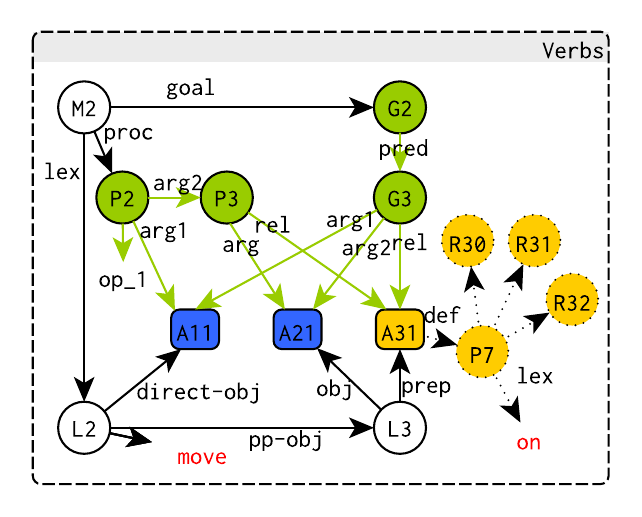}
        \end{subfigure}
    \hfill
    \begin{subfigure}[b]{0.48\textwidth}
    \includegraphics[width=\textwidth]{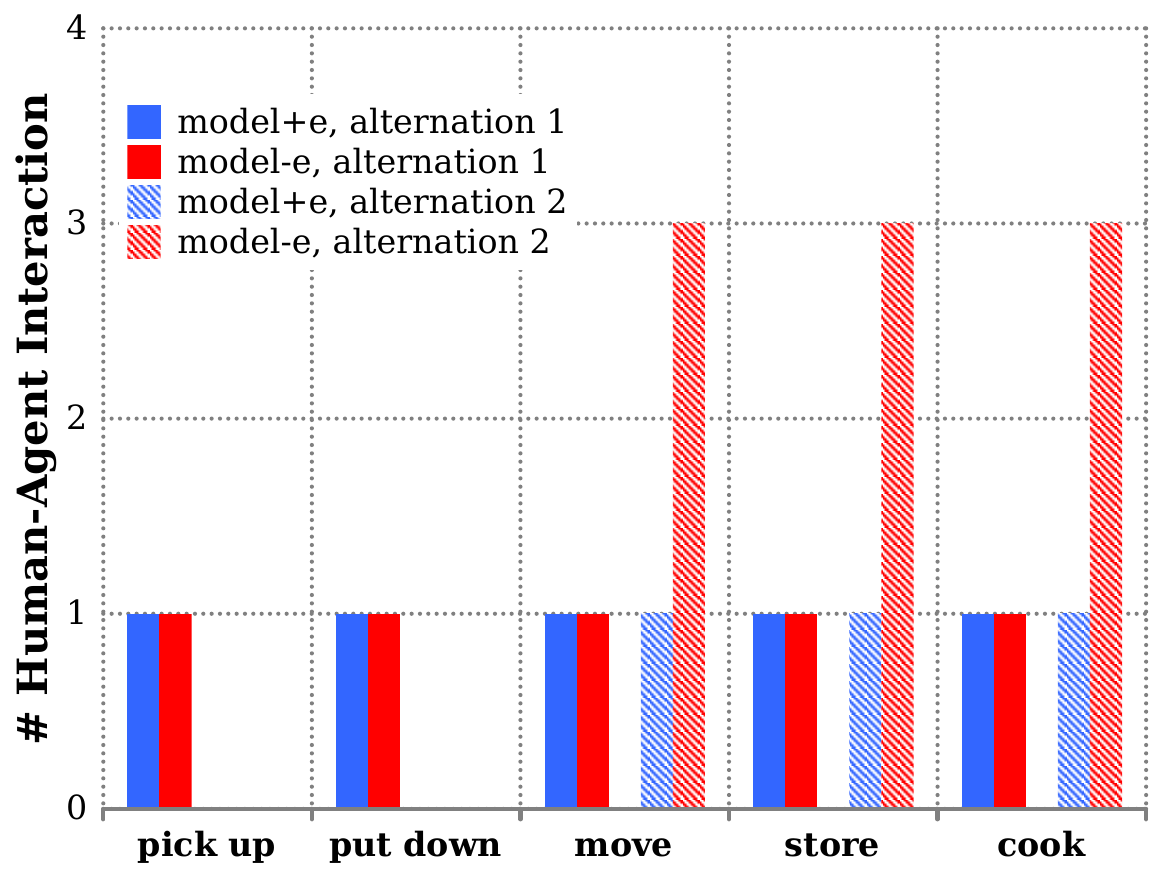}
    \label{fig:missing-object-graph}
    \end{subfigure}
    \caption{(left) Declarative knowledge for \emph{move} after the
      training episode. (right) Number of interactions required for
      comprehending verbs in different alternations.}
    \label{fig:rep}
\vskip -0.2in
\end{figure}

The default configuration can be extracted from the experience of
learning how to perform the \emph{move} task. When Rosie is asked to
execute a task for the first time, it leads the instructor through a
series of interactions to learn the structure of the task. Suppose
assume that Rosie does not know how to perform \emph{move}. On
receiving the imperative sentence (a), it asks a question about the
goal (\emph{what is the goal of the task?}) and the human instructor
replies, \emph{the goal is the green object to the right of the
  table}. By analyzing the sentence and the goal description, Rosie
extracts a general schema that relates the linguistic structure of the
utterance to the goal of the task. It uses a simple heuristic that
information (object, location, and spatial relationship) specified in the
imperative sentence can be generalized away in the goal
definition. The agent assumes that future instances of the verb
\emph{move} will completely specify the goal. At a later stage, Rosie
receives the sentence (b). Using its knowledge of the goal definition,
Rosie attempts to generate an instantiation. This fails because no
relationship is specified. So, Rosie asks the instructor to describe
the goal. The instructor may reply with \emph{the goal is the green
  object is on the table}. By comparing the current situation (for
sentence (b)) and its experience with sentence (a), Rosie concludes
that the verb \emph{move} may be used in two alternations. The
representation of \emph{move} is augmented to reflect that, if the
relationship is not specified, it should attempt to establish the
\emph{on} relationship between the object and the location. Figure
\ref{fig:rep} shows this augmentation to the task-concept network of
the verb \emph{move} as dotted edges and nodes. Note that this network
is an augmented version of the network C in Figure
\ref{fig:indexical-maps}.

When comprehending the verb move in the future, the model can use the
default values to complete the argumentation of the action if those
values are not specified in the linguistic input itself. This lets
the model use Rosie's instructional experience to fill in
information that is not specified in the linguistic input but is
essential for action.

\subsubsection{Evaluation and Analysis}
\emph{\textbf{Experiment}}: In an environment with four objects, we
instructed Rosie to perform eight instances of five tasks using an
uniform distribution over alternations of the relevant verb. Here we
characterize the verbs used in the experiments.
\begin{itemize}
\item The verb \emph{pick} takes a direct object and does not have any
  alternation. Example: \emph{pick up the red cylinder}.
\item The verb \emph{put} takes a direct object and a
  prepositional object and does not have any alternation. Example
  \emph{put down the red cylinder on the table}.
\item The verb \emph{move} has two alternations. The first specifies
  the object, the prepositional object, and the intended spatial relationship
  (as in \emph{move the red cylinder to the right of the table}). The
  second does not specify the spatial relationship between the direct
  and prepositional object (as in \emph{move the red object to the
    table}).
\item The verb \emph{store} has two alternations. The first specifies
  the direct object, the prepositional object, and the intended spatial
  relationship between them (as in \emph{store the red cylinder in the
    pantry}). The second leaves the prepositional object unexpressed
  (as in \emph{store the red cylinder}).
\item The verb \emph{cook} has two alternations. One specifies the
  instrument used for cooking along with the object to be cooked (as in
  \emph{cook the steak on the stove}). The other leaves the instrument
  unexpressed (as in \emph{cook the steak}).
\end{itemize}
The first two verbs are primitives that have been pre-encoded in
Rosie; the last three are acquired through human-agent linguistic
interaction. For training, Rosie was taught the task with the first
alternation of the corresponding verb. After it successfully learns
the task, we asked it to perform the task using the second
alternation. Any questions asked by Rosie during this training episode
were appropriately answered. Two variations of the comprehension model
were evaluated. Model+e uses Rosie's instructional experience to
augment the linguistic input that is missing information required for
task execution. Model-e is a lesioned version of model+e that does not
exploit the instructional experience but relies on asking the
instructor a question for the missing information. Both models were
given the same instructional experience (12 interactions for
\emph{move} and 16 interactions for \emph{cook}). \looseness=-1

\noindent
\emph{\textbf{Results}}: The graph in Figure \ref{fig:rep} shows the
number of interactions that occurred during the comprehension of task
commands in model+e (in blue) and model-e (in red). The patterned bars
correspond to the first alternation and the plain bars correspond to
the second alternation (if applicable). For verbs without alternations
(\emph{pick} and \emph{put}), both models take equal number of
interactions to execute the task (one per task instance). For verbs
with alternations, the models behave differently for different
alternations. For the first alternation in which all information is
specified, both models take one interaction per task. However, for the
second alternation that leaves some argumentation unexpressed, model+e
takes only one interaction per task for performance because it uses
the knowledge acquired through learning to fill in the missing
information. Model-e must ask questions to gather the information
missing from the sentences with unexpressed verb argumentation,
resulting in more human-agent interactions (three per task
instance). Both models comprehend both alternations of verbs and
correctly execute the task.

\section{Related Work and Qualitative Analysis}
\label{related-work}
This paper addresses the challenge of situated language comprehension
for intelligent agents that continually learn from their experience in
complex domains . A primary purpose of a linguistic faculty in an
intelligent agent is the exchange of information with collaborators,
which influences and guides reasoning, behavior, and learning. This
motivates the study of language as a communication system that
functions with, and is informed by, other cognitive capabilities. In
this light, we propose five properties that a situated comprehension
model should incorporate:
\begin{enumerate}[leftmargin=0.8cm]
\item[D1] \emph{Referential}. It must implement a theory of translating
  amodal linguistic symbols to modal representations of beliefs,
  knowledge, and experience that are external to the linguistic
  system. The Indexical Model addresses this by formulating the
  problem of language comprehension as search over perceptions,
  short-term memory, and long-term knowledge. 
\item[D2] \emph{Integrative}. Human language is highly contextual and
  relies on several non-linguistic sources to convey meaning. To
  successful comprehend language, a model must exploit multiple
  information sources, including perceptions, domain knowledge,
  common-sensical knowledge, and short- and long-term experiences. It
  should also readily incorporate information from non-verbal
  communication such as gestures and eye gazes. In the Indexical
  Model, the background knowledge constrains and guides search. We
  have shown how diverse kinds of background knowledge can be
  exploited to generate and evaluate interpretations to handle
  ambiguities and missing information.
\item[D3] \emph{Active}. The model should actively use all its
  knowledge and reasoning capabilities to generate and reject
  candidate hypotheses. Such active processing not only informs
  further communication with the collaborator, by the way of
  requesting for clarification and repetition, but may also inform
  knowledge acquisition. The Indexical Model retains information that
  helps to determine the cause of failure and to generate and ask
  questions in case of ambiguity or missing information.
\item[D4] \emph{Adaptive and expandable}. As Rosie gathers more
  experience and knowledge of the environment, its comprehension
  capabilities must scale elegantly and robustly. This desideratum was
  not explicitly pursued in this paper but our previous work
  \citep{Mohan2012} shows that indexical comprehension scales with
  acquisition of perceptual, spatial, and action knowledge.
\item[D5] \emph{Incremental}. A model must build up the meaning
  representation as each word is processed. Likely continuations may
  be inferred from these partial structures informing linguistic
  perception and syntactic ambiguity resolution. Incremental
  processing was not addressed in this paper and will be studied in
  future.
\end{enumerate}

In the remainder of this section, we use these desiderata to analyze
the approaches to representing and using language semantics in various
fields of AI research.

Research on semantics in computational linguistics and natural
language processing can be broadly categorized into three distinct
groups, \emph{formal}, \emph{distributional}, and \emph{grounded}
semantics. While the earlier two approaches have been well studied in
the literature, the last approach has recently gained momentum. The
formal approaches to semantics represent meaning as amodal first-order
logic symbols and statements. Although, this allows for incorporating
extra-linguistic knowledge during comprehension through inference, the
symbols and predicates are not grounded in the physical
world. Distributional semantics usually incorporates linguistic
contexts with no explicit groundings to the observations from the
environment.

Work on grounded semantics can be characterized as an extension of
formal semantics to include state and action information from
environments. Example applications include navigation tasks
\citep{Chen2011} and RoboCup sportscasting \citep{Liang2009}. These
projects have focused on acquisition (D4) of grounded lexicon and
semantic parsers (D1) from an aligned corpora of agent behavior and
the text that describes it. There are several reasons for why such
approaches cannot be used to design collaborative agents that engage
in situated communication. These methods focus on statistical batch
learning paradigms. Although, this results in comprehension models
robust to errors in linguistic input, they cannot be extended
online. The inability to comprehend an utterance is reported as a
failure and which is not sufficient to drive situated communication or
learning (D3). Further, the work proposes that the entire complexity
of language comprehension can be encoded in a semantic parser and does
not address the use of reasoning mechanisms and background conceptual
knowledge for the purposes of language comprehension. Finally, these
approaches assume a fairly simplistic agent with propositional state
and action representations. This simplistic representation of the
world state and dynamics poses problems in adapting the comprehension
model to agents embedded in physical environments that require
complex, relational representations for reasoning and action. These
approaches do not provide insights on the role of non-linguistic
context on language processing (D5).

In the robotics community, grounded comprehension has been studied in
the context of describing a visual scene \citep{Roy2002}, understanding
descriptions of a scene \citep{Gorniak2004}, understanding spatial
directions \citep{Kollar2010}, and understanding natural language
commands for navigation \citep{Tellex2011}. These comprehension models
work with the complex state and action representations required for
reasoning about physical worlds (D1). Their primary focus has been on
the acquisition of grounding models through batch-learning from
human-generated descriptions of robot's perceptions or
behavior. However, generating an annotated corpus is expensive. The
agents are prone to failure if their training is insufficient for
grounding a novel instruction. An interactive agent on the other hand
will switch to learning mode if it is unable to comprehend the
instruction. It is unclear if such data extensive, corpus-based,
batch-learning mechanisms can be effectively incorporated in online
and incremental human-agent interactions, allowing the agent to guide
communication. Additionally, these mechanism do not address the
challenges arising from ambiguities in natural language.

SHRDLU \citep{Winograd1972} is a well-known early attempt to design an
intelligent agent that could understand and generate natural language
referring to objects and actions in a simple virtual blocks world
(D1). It performed semantic interpretation by attaching short
procedures to lexical units. It demonstrated simple learning as the
user could define compositions of blocks (such as a tower) that the
system would remember and could construct and answer questions about
(D3). The system was not physically grounded, did not learn new
procedures (D4) and, therefore, was constrained to pre-programmed
behaviors.

Ongoing work on Direct Memory Access Parsing (DMAP; Livingston and
Riesbeck, 2009) studies the utility of incorporating information from
ontological and instance-based inference for linguistic processing in
the context of learning by reading. DMAP incrementally integrates
memory during parsing, which can reduce the number of ambiguous
interpretations and reference resolution. DMAP has several desirable
properties. It is referential (D1), integrative (D2), and active (D3)
but it has not been investigated in human-agent interaction
contexts. Our work provides further support to the primary thesis that
linguistic features are cues to the hearer/reader to search their
knowledge and experience.

Other cognitive system research has addressed the challenge of
situated language processing for human-agent
interaction. \citet{Scheutz2004} present a visually-grounded,
filter-based model for reference resolution that is implemented on a
robot with audio and video inputs. Ambiguities are resolved by
accounting for attentional context arising from fixations in the work
area. In related work, \citet{Kruijff2007} demonstrated incremental
parsing at multiple levels that includes non-linguistic contexts, such
as the ongoing dialog and declarative pre-encoded selectional
restrictions along with visual semantics. Apart from being referential
and integrative, these projects address issues that arise in spoken
dialog processing and online, incremental comprehension
(D5). \citet{Brenner2007} describe how action commands can be
interpreted in a task-oriented fashion to identify and instantiate
goals and plans. This model brings in knowledge about initial state
and goal descriptions that are relevant to generating and executing a
plan. This is similar to our indexical comprehension of verbs. Our
work can be viewed as an extension of these efforts to develop a more
complete comprehension model for intelligent agents.

\section{Concluding Remarks}
\label{conclusion}
In comparison to standard approaches to semantics and meaning
representations prevalent in natural language community, the Indexical
approach to language comprehension described in this paper affords
several advantages. Previous approaches either encode semantics as
amodal symbols that that are not grounded in real-wold experiences or
as propositions which do not capture the relational or probabilistic
properties of complex environments. In the Indexical approach,
semantics can be encoded using diverse, modality-specific
representations. These include probabilistic representation for
perceptions, relational representation for spatial reasoning, and
hierarchical policies for task execution, and models for reasoning
about the environmental dynamics. Such representations are typical of
agents designed to function in complex environments. Additionally,
standard learning algorithms (kNN, explanation-based learning) can be
used to expand the agent's knowledge, thereby, extending its situated
comprehension capabilities.

In the formulation of comprehension as a search over short-term and
long-term experiential knowledge, non-linguistic context has a natural
role. It provides constraints over the hypothesis space and guides
search. Non-linguistic context can be derived from various sources
including the ongoing discourse, the current perceptual state, the
knowledge of tasks, and the models of environmental dynamics. Other
cognitive mechanisms such as reasoning and attention also contribute
to comprehension by providing additional constraints on
interpretations. We show that exploiting different contexts in our
model reduces ambiguity in referring expression
resolution. Experiential knowledge augments the linguistic input by
incorporating knowledge from prior experiences with the
environment. This is useful in situations where the linguistic input,
such as in \emph{take the trash out}, is underspecific and does not
encode enough information for reasoning and action.

The focus of our future work will be on studying other linguistic
ambiguities that may arise in instructional interactions and how they
can be addressed by incorporating information from various cognitive
modules. A concern is that one verb word may indicate different task
goals and policies. For example, the sentences \emph{store the rice}
and \emph{store the milk} indicate different goal locations
(\emph{pantry} for \emph{rice} and \emph{refrigerator} for
\emph{milk}). The comprehension model should be able to use the
semantic categorization of the arguments to instantiate the goal with
appropriate locations. Other ambiguities arise in determining the site
of preposition phrase attachment. In the sentence \emph{store the red
  cylinder on the green block in the pantry}, it is unclear if the
phrase \emph{in the pantry} attaches to the verb \emph{store}
directly, or to the phrase \emph{on the green block}. This can be
resolved by the incorporating the current state of the
environment. Another concern is that the described model does not
support interpretation of quantification (\emph{store all red
  objects}) or of categories of objects (\emph{chess pawns cannot not
  move backwards}). This limits what can be expressed in instructions
for tasks and games. A different dimension of future research is
incremental comprehension that would lead to robust performance on
incomplete and ungrammatical linguistic input.

This paper has focused on mechanisms that are useful in language
comprehension where the elements of an utterance can be directly
grounded in the shared state between the speaker and the
hearer. However, much of human communication is non-situated where
people talk about scenarios that cannot be directly perceived and may
have occurred in the past (retrospective) or is expected to occur in
the future (prospective). Human hearers readily generate
perceptual simulations guided by the content of the utterance and can
use these models for reasoning about the scenario being described. The
perceptual simulations are informed by the hearer's experience of the
world. In future, we are interested in expanding the indexical
approach described to address non-situated comprehension and its
interaction with knowledge acquisition and learning.

\section*{Acknowledgments} 
The authors acknowledge the funding support of the Office of Naval
Research under Grant Number N00014-08-1-0099. The authors thank Edwin
Olson and the APRIL lab at the University of Michigan, Ann Arbor, for
the design and implementation of the robot including its perception
and actuation algorithms. The authors also appreciate the insightful
comments and suggestions made by the reviewers.

{\parindent -10pt\leftskip 10pt\noindent
\printbibliography}

\end{document}